\PassOptionsToPackage{prologue,usenames,dvipsnames}{xcolor}
\documentclass{article}

\usepackage{arxiv}
\usepackage{amsmath,amsfonts}
\usepackage{algorithmic}
\usepackage{graphicx}
\usepackage{changepage}
\usepackage{textcomp}
\usepackage{xspace,xpunctuate}
\usepackage{array}
\usepackage{tabularx}
\usepackage{enumitem}
\usepackage{seqsplit}
\usepackage{pgf}
\usepackage{rotating}
\usepackage{multicol}
\usepackage{multirow}
\usepackage{authblk}
\usepackage{hyperref}
\usepackage{footmisc}
\usepackage{natbib}
\usepackage{booktabs}

\hypersetup{
    colorlinks=true,
    linkcolor=red,
    urlcolor=blue,
    citecolor=teal
}
\let\oldfootnote\footnoterule
\renewcommand*\footnoterule{\vfill\oldfootnote}

\title{Multimodal Fusion with LLMs for Engagement Prediction in Natural Conversation}
\renewcommand{\undertitle}{\vspace{-4\parskip}}
\renewcommand{\headeright}{}
\date{}

\newbox{\orcid}
\sbox{\orcid}{\includegraphics[scale=0.06]{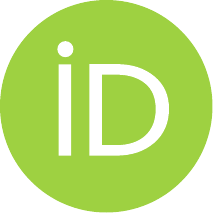}}

\makeatletter
\let\oldauthor\@author
\renewcommand{\@author}{\vskip -2.5\parskip \oldauthor}
\makeatother

\newcommand{\addauthor}[3]{%
\author[#1]{%
        \ifx&#3& 
            \raisebox{0pt}[0pt][0pt]{#2}%
        \else%
            \href{#3}{\raisebox{0pt}[0pt][0pt]{\usebox{\orcid}\hspace{1mm}#2}}%
        \fi%
    }%
}

\begin{document}

\newcommand{\CMU}{Carnegie Mellon University, Pittsburgh, PA 15213}

\addauthor%
    {1}{Cheng Charles Ma\footnote[1]{}\footnote[2]{}}{https://orcid.org/0009-0004-4036-0390}
    
\addauthor%
    {2}{Kevin Hyekang Joo\footnote[1]{}\footnote[3]{}}{https://orcid.org/0000-0002-6387-5686}

\addauthor%
    {2}{Alexandria K. Vail\footnote[4]{}}{https://orcid.org/0000-0001-5221-4092}

\addauthor%
    {3}{Sunreeta Bhattacharya}{https://orcid.org/0000-0002-6292-6699}
    
\addauthor%
    {2}{\'{A}lvaro Fern\'{a}ndez Garc\'{i}a}{https://orcid.org/0009-0001-1664-9272}

\addauthor%
    {1}{Kailana Baker-Matsuoka}{https://orcid.org/0009-0001-6801-253X}

\addauthor%
    {1}{Sheryl Mathew}{https://orcid.org/0009-0001-6549-9418}

\addauthor%
    {4,5}{Lori L. Holt}{https://orcid.org/0000-0002-8732-4977}

\addauthor%
    {2}{Fernando De la Torre}{https://orcid.org/0000-0002-7086-8572}

\affil[1]{Computer Science Department, \CMU}
\affil[2]{Robotics Institute, \CMU}
\affil[3]{Neuroscience Institute, \CMU}
\affil[4]{Department of Psychology, The University of Texas at Austin, Austin, TX 78712}
\affil[5]{Center for Perceptual Systems, The University of Texas at Austin, Austin, TX 78712}

\newcommand{\emailcharles}{ccma@cs.cmu.edu}
\newcommand{\emailkevin}{khjoo@usc.edu}
\newcommand{\emailalex}{avail@cs.cmu.edu}

\newcommand{\emaillink}[1]{\texttt{\href{mailto:#1}{#1}}}

\newcommand\blfootnote[1]{%
  \begingroup
  \renewcommand\thefootnote{}\footnotetext{#1}%
  \endgroup
}

\blfootnote{\footnotesize{
\textasteriskcentered\ (equal contribution)
\quad \textdagger\ \emaillink{\emailcharles} 
\quad\textdaggerdbl\ \emaillink{\emailkevin} 
\quad\textsection\ \emaillink{\emailalex}}
}

\maketitle

\begin{figure*}[h]
 \centering\vspace{-4em}
 \includegraphics[width=\textwidth]{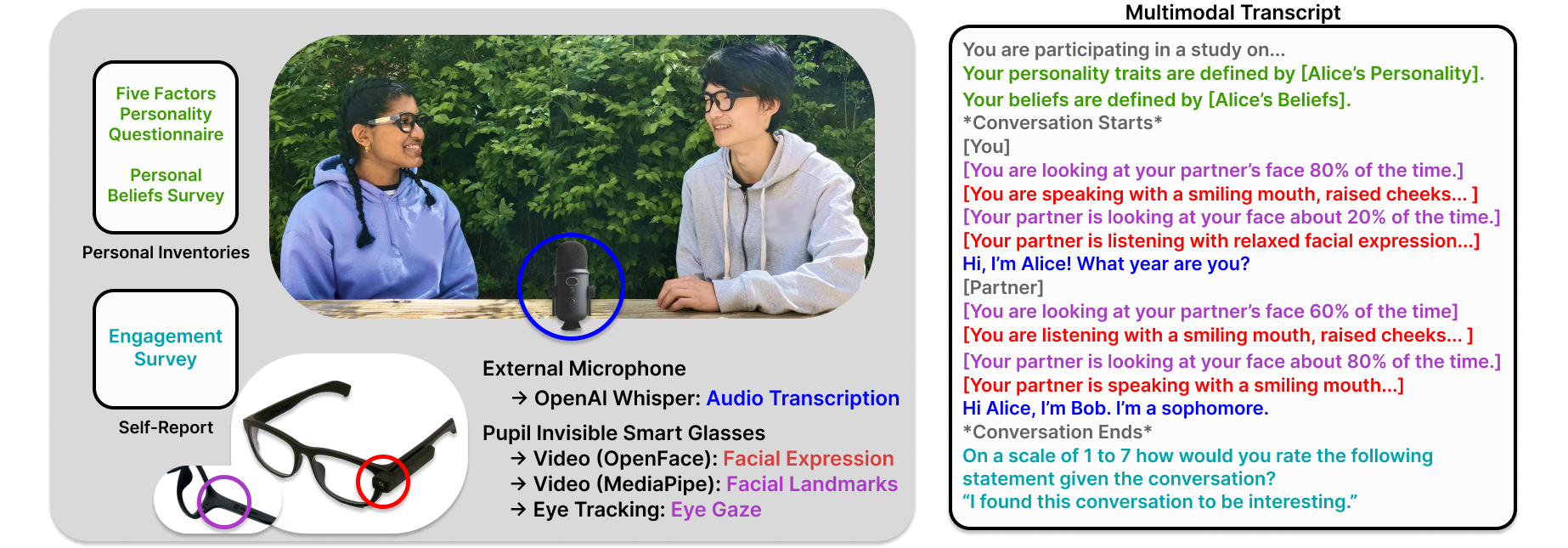}
 \caption{\small{Visual representation of recorded behavior modalities during casual conversation and a sample of the multimodal transcript illustrating their fusion as introduced in this work. The goal is to predict engagement from this multimodal data. Color-coded modality names correspond to lines of the same color in the multimodal transcript.\label{fig:teaser_figure}}
 }
\end{figure*}

\begin{abstract}
    \small{Over the past decade, wearable computing devices (``smart glasses'') have undergone remarkable advancements in sensor technology, design, and processing power, ushering in a new era of opportunity for high-density human behavior data. Equipped with wearable cameras, these glasses offer a unique opportunity to analyze non-verbal behavior in natural settings as individuals interact. Our focus lies in predicting engagement in dyadic interactions by scrutinizing verbal and non-verbal cues, aiming to detect signs of disinterest or confusion. Leveraging such analyses may revolutionize our understanding of human communication, foster more effective collaboration in professional environments, provide better mental health support through empathetic virtual interactions, and enhance accessibility for those with communication barriers.

In this work, we collect a dataset featuring 34 participants engaged in casual dyadic conversations, each providing self-reported engagement ratings at the end of each conversation.
We introduce a novel fusion strategy using Large Language Models (LLMs) to integrate multiple behavior modalities into a ``multimodal transcript'' that can be processed by an LLM for behavioral reasoning tasks. Remarkably, this method achieves performance comparable to established fusion techniques even in its preliminary implementation, indicating strong potential for further research and optimization. 
This fusion method is one of the first to approach ``reasoning'' about real-world human behavior through a language model. 
Smart glasses provide us the ability to unobtrusively gather high-density multimodal data on human behavior, paving the way for new approaches to understanding and improving human communication with the potential for important societal benefits. The features and data collected during the studies will be made publicly available to promote further research.

}
\end{abstract}

\section{Introduction}

Wearable computing devices, also known as ``smart glasses,'' offer new approaches to quantifying and understanding human behavior through unobtrusive, high-density behavior tracking. Equipped with sensors such as a video scene camera to monitor the wearer’s view, an eye camera to estimate gaze, a microphone to record speech, and an inertial measurement unit to measure head orientation, smart glasses can capture and respond to human behavior as it unfolds in real-time and real-world contexts. There are numerous potential future applications for such systems: for example, facilitating navigation among the visually impaired, or augmenting social cues for individuals with difficulties reading nonverbal signals.

Although there has been substantial prior research in laboratory settings~(\cite{cafaro2017, ringeval2013, vinciarelli2009a}) and human-agent interaction~(\cite{ben-youssef2017, celiktutan2019, mckeown2012}), there are still many rich, unexplored opportunities in natural social contexts, for which smart glasses offer unique capabilities for study. With smart glasses, we can capture natural social interactions that are not constrained by the artificial settings of a laboratory, but rather occur in the natural course of daily life, as we seek help, share information, learn, and maintain social bonds through face-to-face communication. These interactions are rich, nuanced, and impacted moment-by-moment by multimodal cues, both overt and subtle. The stakes can be high: human conflict --- between couples, among friends and families, in leadership and governing bodies, and even among societies --- occurs when communication breaks down. Face-to-face communication is fundamental in maintaining group cohesion, preserving mental health, fostering academic learning, and supporting developmental growth.

Engagement has long been recognized as a key determinant of communication success. While lacking a precise definition, engagement can be loosely defined as an individual's attentional and emotional investment during communication~(\cite{pellet-rostaing2023}). The ability to captivate in conversation can determine life-changing interactions, whether acing a job interview or making a favorable impression on a first date. The depth of our engagement and that of our partner shapes the outcomes of many social, educational, and professional activities.

For the most part, humans automatically and implicitly pick up on the subtle, variable cues that convey engagement in a conversation. Yet, building systems that accurately measure and gauge conversational engagement remains a formidable challenge. Difficulties arise with the complexity and subtlety of human behavior, its context-dependence, and its variability across personal histories and cultural backgrounds. Further complicating matters, social communication is inherently multifaceted, with engagement likely to be conveyed across verbal content, nonverbal cues like tone of voice, facial expressions, hand and head gestures, and also through the absence of overt signals, such as extended periods of silence or few gazes to a partner’s face. The unpredictable and dynamic nature of social exchanges makes predicting engagement difficult, as engagement patterns can shift rapidly and vary widely across contexts. Thus, techniques that can perform effectively with minimal or no in-domain training are of particular interest.

The dearth of relevant data presents another challenge. Although there is an abundance of openly available datasets of dyadic interactions from a third-person viewpoint, such as IEMOCAP~(\cite{busso2008}), SEMAINE~(\cite{mckeown2012}), MEISD~(\cite{firdaus2020}), MELD~(\cite{poria2019}), or NoXi~(\cite{cafaro2017}), naturalistic dyadic interactions captured from an egocentric viewpoint are scarce. In the past few years, as smart glasses have become more widely accessible, research has begun to gather egocentric recordings for other tasks, such as skilled human activity (Ego-Exo4D,~\cite{grauman2024}) and user gaze anticipation~(\cite{lai2023listen}), though less focused on interpersonal behavior. These factors pose significant challenges for building socially-aware artificial systems that accurately interpret and respond in a manner that feels authentic and engaging to humans. Nonetheless, there is good reason to work to meet these challenges. Imagine a system that can gauge audience engagement with a teacher’s lecture and provide on-the-fly feedback they can use to better engage their students. Or consider assistive technologies that can offer alternative presentations of challenging social signals for those with communication disorders. The potential applications are extensive.

The contribution of the present work is twofold. We introduce a novel dataset including recordings of natural, unscripted conversations among unfamiliar dyads wearing Pupil Invisible smart glasses, as illustrated in the left-hand segment of \autoref{fig:teaser_figure}. This dataset contains conversations between 19 unique dyads, including video and audio recordings, eye tracking, and self-reported information on demographic, political, and personality factors from the participants.

The second contribution presents an analysis of this dataset, focusing on predicting participant engagement levels through post-session self-reports. We compare audio-visual classical fusion techniques~(\cite{xu2022, wuerkaixi2022}) with our novel proposed fusion approach, which uses a large language model (LLM) as a reasoning engine to fuse behavioral measures into a multimodal textual representation, a sample of which is displayed in the right-hand segment of \autoref{fig:teaser_figure}. Our results indicate that this approach achieves performance comparable to the established fusion techniques even in this early implementation. This approach is a powerful, simple, and flexible framework for future work on modeling human behavior and developing socially intelligent technologies.

\section{Prior Work}

This section reviews classical fusion methods and works on LLMs for analyzing and understanding human behavior.

\subsection{Classical Fusion}
\citeauthor{curhan2007} used speech features (conversational engagement, prosodic emphasis, and vocal mirroring) in the first five minutes of a simulated negotiation to predict the outcomes of the negotiation~(\cite{curhan2007}). Using these features, they predicted 30\% of the variance in negotiation outcomes, demonstrating the value of speech features in conversational dynamics. This result suggests that speech features are similarly important in predicting conversational engagement. Activity level and mirroring had differing relationships with the outcome depending on the assigned position of participants, showing that perceived status can affect how conversational dynamics relate to negotiation success. This interaction poses the question of how status affects how features predict conversational engagement. 

\citeauthor{pellet-rostaing2023} used prosodic-acoustic, prosodic-temporal, mimo-gestural, and linguistic features to predict the engagement level of the target participant while holding the speaking turn~(\cite{pellet-rostaing2023}). The study showed the value of visual and audio features, achieving the best results with the prosodic-acoustic, prosodic-temporal, and mimo-gestural modalities. Achieving similar results to studies using annotator-defined segments demonstrated that annotating engagement at a turn level can be effective. 

In our study, we attempted to use gaze as a means of gauging dyadic interaction, along with other modalities, as it is evidenced by some to have correlations with engagement~(\cite{goodwin1981, ranti2020}). \citeauthor{goodwin1981} emphasizes the interconnected nature of gaze behavior among participants in a conversation and points out that the way individuals direct their gaze is not a solitary or random act but is deeply intertwined with the social dynamics of the interaction~(\cite{goodwin1981}). This gaze behavior acts as a nuanced signal of a participant's level of attention and engagement, reflecting whether they are actively participating or disengaging from the conversation. Furthermore, \citeauthor{goodwin1981} explores the concept of gaze withdrawal as a strategic communicative gesture that participants use to signal their intentions within the conversation, such as making a bid for closure or expressing a particular understanding of the conversation's trajectory. 

Moreover, \citeauthor{ranti2020} underscore the potential of utilizing eye-blink measures as a reliable indicator of an individual's subjective engagement with various stimuli~(\cite{ranti2020}). By closely analyzing the timing of blink inhibition in response to unfolding scene content, they found that they could uncover the viewers' unconscious, subjective evaluations of the importance and engagement level of what they observe. A notable observation is that a slower blinking rate is often associated with a higher degree of engagement, suggesting that individuals are more absorbed and attentive to the conversation or content presented to them.

\subsection{Large Language Models (LLMs)}
The capabilities and accessibility of LLMs have opened up a wide range of potential applications, particularly in fields related to human subjects like psychology. They range from creating synthetic datasets of LLM-generated responses in human-less experiments~(\cite{demszky2023}) to providing automated feedback to clinicians~(\cite{stade2024}).

One application involves exploring the ability of LLMs to mimic human behavior because of their potential to reduce the need for human subject experiments and power realistic, interactive interactions. \citeauthor{aher2023} explore the ability of LLMs to reproduce human subjects' behavior in classic experiments, such as the ``Wisdom of Crowds''~(\cite{aher2023}). \citeauthor{argyle2023} investigate the potential of LLMs as proxies for human sub-populations in social science research~(\cite{argyle2023}). \citeauthor{tavast2022language} evaluate the human-likeness of responses on the PANAS questionnaire generated by \texttt{GPT-3} ~(\cite{tavast2022language}). The feasibility of using LLMs to replace human participants is further explored in ~(\cite{harding2023ai, dillion2023can}). \citeauthor{park2023} introduce generative agents powered by LLMs that simulate believable human behavior in a virtual environment~(\cite{park2023}), also similarly seen in \cite{zhou2024}. There is also a body of work on understanding the personality of LLMs, identifying ways to manipulate the personality embodied by an LLM, and injecting personality into LLMs to predict human responses concerning values~(\cite{serapio-garcia2023, kang2023}). 

Another application involves exploring the ability of LLMs to understand human behavior. This line of work involves evaluating their theory of mind abilities, which refers to the ability to understand the mental states of others, such as purpose or intention ~(\cite{premack1978}). Prior work has proposed various benchmarks and methods to evaluate an agent's theory of mind~(\cite{kim2023, sap2019, sap2022}). 

These works are essential to assessing the ability of LLMs to simulate and understand human behavior. However, they are all limited to static benchmarks or simplified virtual interactions. There is a lack of work exploring the ability of LLMs to simulate and predict the outcomes of human social interactions, such as predicting a person's responses to a survey that measures engagement. We argue that this dimension should be considered when developing LLMs to simulate and understand behavior.

Our work proposes a dataset and method for unifying the work on simulating and understanding engagement in social interactions with LLMs grounded in real, in-the-wild social interactions. Given the potential of LLMs to advance socially intelligent technologies, incorporating in-the-wild social interactions into research is essential. 

\section{Data Set}

In our work, we collected and studied pairs of strangers conversing in a room, recorded from the viewpoint of each participant through a pair of smart glasses. 

\subsection{Population}

Our study contained 34 unique participants and 19 unique dyads\footnote{Two participants appeared in multiple dyads, but all dyads were unique.}. Demographically, 14 participants identified as male, 19 identified as female, and one identified as non-binary; 47\% identified as Asian, and 38\% identified as White/Caucasian. All participants were 18--35 years of age but were primarily in their early twenties. Participants were recruited from the local university through various physical and digital media and word-of-mouth. Participants were required to be fluent in English and have normal or corrected vision with contact lenses (to avoid conflict with the smart glasses).  

\subsection{Procedure}

The entire recording session lasted approximately 15 minutes, including introductions and closing. Each participant was equipped with a pair of smart glasses (refer to \autoref{sec:recording} for detailed specifications), capturing their field of vision, head motion, and gaze. While the smart glasses are advertised to work well across recording sessions without calibration, they benefit from calibration when changing users~(\cite{tonsen2020})\footnote{Note that the referenced study was conducted by the manufacturer of the device.}, so we conducted a calibration procedure for each participant before the beginning of the session. At the start of the session, participants were suggested an ice-breaker: their experience during COVID-19, a universally shared topic, but the conversation was not constrained to this topic. Following the session, participants completed questionnaires on their beliefs, personality, and engagement during this interaction (refer to \autoref{sec:questionnaires} for details on the questionnaires).

\subsection{Recording Instruments\label{sec:recording}}

Each session was recorded using Pupil Invisible smart glasses worn by all participants and a centrally placed external microphone to record the dialogue.

\subsubsection{Pupil Smart Glasses\label{sec:glasses}}

Each participant was equipped with Pupil Invisible\footnote{\url{https://pupil-labs.com/products/invisible}} smart glasses manufactured by Pupil Labs, specially designed to closely resemble regular eyeglasses for user comfort and a discreet appearance. The key features of these smart glasses that we leverage in our work include the following:

\begin{itemize}
    \item \textbf{Scene camera:} A detachable camera mounted on the left arm of the glasses frame captures the wearer's field of view with an $82^\circ \times 82^\circ$ viewing angle, at a resolution of $1088 \times 1080$ pixels and a frame rate of 30 Hz.
    \item \textbf{Eye gaze tracking:} Two IR cameras, positioned near the would-be hinges of the glasses frame, record eye movements at a resolution of $192 \times 192$ pixels and a frame rate of 200 Hz. Post-processing software provided by the manufacturer converts this data into 2D gaze points at 120 Hz in scene camera coordinates. This system is advertised to achieve an uncalibrated accuracy of approximately $4.6^\circ$, but calibration per user can enhance accuracy~(\cite{tonsen2020}).
\end{itemize}

\subsubsection{Stereo Microphone}

In addition to the recordings captured by the smart glasses' scene camera, we used an external high-quality stereo microphone (Zoom H4N Pro) to record the conversation at a standard 44.1 kHz sampling rate. This decision was made after determining that the quality of the audio captured by the smart glasses scene camera was insufficient for acoustic analysis. To synchronize the media streams, participants were instructed to perform a hand clap at the start of each session, emulating the clapperboard technique commonly used in film production.

\subsection{Self-Report Questionnaires\label{sec:questionnaires}}

The participants were asked to complete a questionnaire that measured self-reported engagement after each interaction. The engagement questionnaire consisted of 53 items based primarily on previous studies on participant perception of interaction quality~(\cite{cuperman2009}): refer to \autoref{appendix:questionnaire} for detailed information and statistics regarding the items in the engagement questionnaire. The participants were also asked to complete the Big Five Inventory~(\cite{mccrae1999}) for personality information and a handcrafted questionnaire on personal beliefs. This questionnaire was based on a set of socio-cultural issues studied to gauge polarization along the political spectrum~(\cite{pewpolitical2021}).

\section{Feature Extraction\label{section:feature_extraction}}

After gathering the recordings, the data required pre-processing. Initially, we adjusted the video to eliminate the radial distortion introduced by the scene camera's lens. This was achieved by applying the distortion coefficients provided by the manufacturer\footnote{For details on the information provided by the manufacturer, see Pupil Lab Invisible recording and export instructions: \url{https://docs.pupil-labs.com/invisible/data-collection/data-format/}.}. Due to the differing frame rates between the eye-tracking camera and the egocentric scene-view camera, we also synchronized the data to a unified 30 fps timestamp. 

\subsection{Facial Expression}

Facial action units (FAU) from the processed video were extracted with OpenFace 2.0~(\cite{baltrusaitis2018}). Since OpenFace achieves optimal performance when the face in the image exceeds a width of 100px, we needed to upscale our data to meet this suggestion. For each frame, we used MediaPipe~(\cite{lugaresi2019})\footnote{MediaPipe version: 0.9.1} to identify the location of the face in the image, then cropped and rescaled the image to ensure that the face was centered and was at least 240px wide and the final dimensions were 1080x1080px. If no face was detected for a particular frame, the location of the face in the previous frame was used. 

\subsection{Gaze Tracking}

For every frame, we determined whether a participant's gaze is directed towards their partner's face, recognizing the significance of gaze in forecasting engagement~(\cite{celepkolu2018predicting, nakano2010estimating}).
This was accomplished by creating a convex hull using the 478 2-dimensional face landmarks extracted from MediaPipe to outline the face. A gaze point captured by Pupil smart glasses was deemed to be on the face if it fell within the convex hull (including its boundary) or within 30\% of the width of the face's convex hull to account for the potential inaccuracy of gaze prediction from Pupil smart glasses described in its specifications.

\subsection{Dialogue Transcription\label{sec:transcription}}
OpenAI's Whisper~(\cite{radford2023})\footnote{Whisper version: large-v20230918} was used to transcribe the recording from each session. Whisper outputs fine-grained segments with start and stop times around a few seconds long. A speaker was assigned to each segment. If the segment contained speech from both speakers, the speaker who spoke most was assigned. Diarization tools like PyAnnote~(\cite{Plaquet23, bredin2023}) and source separation tools performed poorly with audio from our dataset, so manual labeling was chosen.

\section{LLM Fusion}
 
In this work, we explore the use of large language models (LLMs) to ``reason'' about a social interaction using multimodal information. Our method involves prompting an LLM to simulate a study participant and answer the end-of-session engagement questionnaire as though it were the participant theirselves.

\subsection{Socratic Models}

Understanding and interpreting the reasoning of machine learning models is widely recognized to be a significant challenge. Typically, models encode behavioral features into a high-dimensional, abstract vector space, which is then mapped onto the target prediction space. To understand a model's inner workings, we usually project these intermediate data into a space that is more understandable to humans, often through visualization techniques. However, consider the possibility of the inverse --- rather than allowing the model to obscure information into abstract dimensions, we could direct its operation into a universally interpretable space: the domain of language itself. When studying a topic like human behavior from a computational perspective, AI systems like LLMs that utilize language to ``reason'' about said topics are worth further study because the language allows for nuance and ambiguity that inherently exists in these fields.

Socratic Models, named for the ancient Greek philosopher's teaching method through cross-examination, use language to integrate information from a diverse set of modalities~(\cite{zeng2022}). Within this framework, pre-trained models fine-tuned toward specific modalities or behaviors translate their interpretations of inputs into natural language. This translation is formulated into a language prompt to direct the reasoning of an LLM. This approach allows a set of pre-trained models to ``discuss'' various multimodal information, akin to asking and answering questions in a Socratic dialogue. By framing the task as a language-driven exchange, the Socratic Model framework allows pre-trained models, each specialized in a distinct domain, to perform downstream multimodal tasks without further training or fine-tuning.

Thus far, there have been only a few early attempts at applying this framework for prediction. In the domain of image captioning, one study revealed that an ensemble of models within the Socratic Models framework generated captions that substantially improve the capabilities of the zero-shot state-of-the-art ZeroCap~(\cite{tewel2022}). However, when compared to fine-tuned models such as ClipCap~(\cite{mokady2021}), performance was not as impressive; yet, this performance gap narrowed considerably when the ensemble was provided a small set of example captions from the training set, suggesting its potential in few-shot learning scenarios~(\cite{zeng2022}).

This concept of ``many-to-one'' alignment has also been explored from other angles. ImageBind, for instance, develops a multimodal representation through a set of image-paired modalities~(\cite{girdhar2023}) while LanguageBind extends video-language pre-training to a broader range of language-paired modalities~(\cite{zhu2024}). However, both of these models still face the challenge of abstracting information. ImageBind and LanguageBind create ``bindings'' centered around a specific modality but do not explicitly work within that modality itself. Instead, they map a primary modality into an abstract space and then align information from other modalities to this space, resulting in a multimodal representation that resembles the embedding of the primary modality. While this approach has proven effective at abstract tasks such as video-text alignment and image-text retrieval, it is less effective in providing human users with a coherent understanding of its reasoning. Our research aims to follow a similar path but with a crucial distinction: our embedding space is designed to be language itself, which may offer a more direct and interpretable framework for multimodal learning. 

Previous studies have established the value of the language modality in understanding complex social phenomena, such as rapport~(\cite{carmody2017}), affinity~(\cite{ireland2010, ireland2011}), and, as in the present work, engagement~(\cite{babcock2014}). Various computational methods have been employed to extract this information from language, from rudimentary bag-of-words approaches to more sophisticated neural network models~(\cite{tausczik2010, vanswol2019}). Recent advancements, however, have seen a considerable increase in LLMs adapted to augment tasks requiring social intelligence: notable applications have included refining persuasive communication for public health campaigns~(\cite{karinshak2023, cox2023}) and identifying adverse social determinants of health within free-form clinical notes ~(\cite{guevara2024}). One of the objectives of the present work is to explore the utility of LLMs for behavior analysis of social interactions: in our case, estimating the conversational engagement of speakers in a dyadic interaction. The proposed approach centers around employing OpenAI's \texttt{GPT} models to impersonate each participant in the conversation by responding to the self-reported questionnaire in a zero-shot manner. This is achieved through reconstructing the conversation using multimodal-informed prompting that combines behavioral information inspired by the Socratic Models framework proposed by \citeauthor{zeng2022}.

\subsection{Algorithms for LLM Fusion}

The novel LLM fusion approach that we introduce enables an LLM to emulate a participant by creating a multimodal prompt: a dialogue transcript of the recording session augmented with textual representations of non-verbal behavior. In this work, these textual representations are formed from the data collected by the smart glasses, multiple pre-trained models, and personality questionnaires, but this method can be extended to contain any number of additional behavioral cues. We aim to evaluate whether this multimodal transcript effectively captures the dynamics of social interaction and can enable an LLM to predict self-reported engagement levels effectively. This work focuses on OpenAI's models \texttt{GPT-4} and \texttt{GPT-3.5} \footnote{Fixed at versions \texttt{GPT-4-0613} and \texttt{GPT-3.5-turbo-0613} for consistency.}, but the technique could be applied to any LLM.

\subsubsection{Modalities\label{text_modalities}}
As described in \autoref{section:feature_extraction}, this analysis included information from speech, gaze, and facial expression modalities, given their straightforward translation into text form and their established significance in signaling engagement.

The \textbf{speech} modality serves as the foundation of the multimodal transcript: its representation consists of the dialogue transcript augmented with speaker-labeled segments as described in \autoref{sec:transcription}.
The \textbf{gaze} modality is represented by a string indicating the proportion of time a speaker's gaze remains on their partner's face, rounded to the nearest 10\% for brevity.

The \textbf{facial expression} modality is represented by a text description of the dominant emotional expression for each speaker-labeled segment of the recording following the methods of existing research~(\cite{tejada2022}) and applications (iMotion's Affectiva;~(\cite{mcduff2016})), these emotional expressions were defined by the facial action units measured by OpenFace 2.0: \emph{happy}, \emph{sad}, \emph{surprise}, \emph{fear}, \emph{anger}, \emph{disgust}, \emph{contempt}, or \emph{neutral}~(\cite{ekman1978}). The \emph{neutral} label was assigned if none of these labels were applicable. The emotional labels were translated into text as described by \citeauthor{zhao2023}, which was generated by prompting ChatGPT, achieving state-of-the-art performance on the Dynamic Facial Expression Recognition problem~(\cite{zhao2023}).

Participant responses to the \textbf{personality} and \textbf{beliefs} questionnaires were also included as part of the \emph{system message}\footnote{\url{https://microsoft.github.io/Workshop-Interact-with-OpenAI-models/Part-2-labs/System-Message}}, providing additional speaker-specific context, as personal characteristics are known to affect a person's social behavior~(\cite{celiktutan2019}).

\subsubsection{Multimodal Transcript Generation}
The messages provided to \texttt{GPT} use the discrete segments in Whisper's transcription as atomic units to which information from other modalities is added. Consecutive segments with the same speaker are merged to combine speech and other modalities into a larger temporal window. 

\texttt{GPT} imitates each participant using the following procedure.  Each merged segment of speech forms the basis of a message provided to OpenAI's ChatCompletion API\footnote{\url{https://platform.openai.com/docs/api-reference/chat/create}}. For each message, the \emph{role} is assigned to the \emph{assistant} if the segment is spoken by the simulated participant or to the \emph{user} if spoken by the partner. The final \emph{user} message is always a questionnaire item introduced by the ``experimenter'' (see \autoref{appendix:questionnaire} for questionnaire details). The final \emph{assistant} message is generated by \texttt{GPT} as a response to the introduced questionnaire item. Prompted transcripts were truncated to five minutes, as previous literature has established that the first five minutes of a conversation is enough information for humans to predict its outcome successfully~(\cite{curhan2007}). This limitation brought the added benefit of reducing the cost of the experiment.

\begin{table*}
\centering\small\makebox[\textwidth]{
\begin{tabular}{l||ccccc|cc}
    & \multicolumn{7}{c}{First-Person Ratings (\autoref{sec:questionnaires}) --- lower RMSE scores are better $\downarrow$} \vspace{0.5em}\\
    \toprule
    Behavior Features   & KNN  & SVM  & RF    & Bi-LSTM   & MLP        & LLM-4  & LLM-4S      \\
    \midrule
Gaze-Only           & 1.556 (0.313)     & \textbf{1.281 (0.310)}     & 1.355 (0.298)     & 1.588 (0.340)     & 1.881 (0.360)     & ---               & ---           \\
Face-Only           & 1.530 (0.322)     & \textbf{1.328 (0.333)}     & 1.390 (0.288)     & 1.563 (0.353)     & 2.090 (0.454)     & ---               & ---           \\
Text-Only           & 1.512 (0.287)     & \textbf{1.280 (0.309)}     & 1.301 (0.287)     & 1.478 (0.358)     & 2.069 (0.432)     & 1.669 (0.396)     & \textbf{1.376 (0.381)} \\ 
Face + Gaze         & 1.500 (0.294)     & \textbf{1.296 (0.339)}     & 1.314 (0.339)     & 1.517 (0.331)     & 1.833 (0.389)     & ---               & ---           \\ 
Text + Gaze         & 1.557 (0.286)     & \textbf{1.291 (0.305)}     & 1.409 (0.289)     & 1.466 (0.307)     & 1.988 (0.405)     & 1.418 (0.394)     & \textbf{1.338 (0.378)} \\ 
Text + Face         & 1.521 (0.289)     & \textbf{1.287 (0.337)}     & 1.305 (0.339)     & 1.572 (0.352)     & 1.945 (0.475)     & 1.477 (0.425)     & \textbf{1.368 (0.417)} \\ 
Text + Face + Gaze  & 1.557 (0.300)     & 1.327 (0.342)     & \textbf{1.303 (1.290)}     & 1.592 (0.355)     & 1.773 (0.356)     & 1.442 (0.423)     & \textbf{1.364 (0.387)} \\ 
    \bottomrule
\end{tabular}}
\caption{\label{tab:rmse_prediction} Prediction performance of classical vs. LLM fusion models when provided data from a limited set of modalities: RMSE mean and standard deviation across validation folds (lower is better). LLM-4/4S refers to ablations with GPT-4.}
\end{table*}

\section{Experiments}
We conducted two experimental series to predict engagement based on the end-of-session questionnaires, outlined in \autoref{sec:questionnaires}. The first series assessed multimodal fusion using classical models (\autoref{section:classical}), while the second series used large language models (LLMs; \autoref{subsect:llmfusion}).
\subsection{Classical Fusion\label{section:classical}}

Five standard machine learning techniques were employed to establish a comparative baseline: $k$-nearest neighbors (KNN), support vector machines (SVM), random forests (RF), bidirectional long short-term memory networks (Bi-LSTM), and multi-layer perceptrons (MLP). Each model was trained using per-turn behavioral features alongside the corresponding self-report ratings for each session, described in \autoref{sec:questionnaires}. The KNN, SVM, and RF models implemented either the multivariate sequence kernel or the global alignment kernel (GAK;~\cite{10.5555/3104482.3104599}) to facilitate the comparison of sequences of varying lengths, as these models are not inherently designed to process sequential or variable-length input. Conversely, the MLP and Bi-LSTM models followed canonical architectures specific to their respective methodologies.

The representations of the behavioral features provided to these models were designed to reflect the information presented to the large language model in \autoref{subsect:llmfusion}. Facial expression was denoted by a label indicating the predominant perceived emotion, while gaze direction was quantified as the proportion of time an individual directed their attention towards their partner's face. These representations parallel the descriptions provided to the LLM via the multimodal transcript. Dialogue text was encoded using sentence embeddings generated through the SimCSE framework (Simple Contrastive Learning of Sentence Embeddings;~\cite{gao-etal-2021-simcse}). For additional information regarding the extraction of these features, refer to \autoref{section:feature_extraction}.

All models were trained and evaluated using leave-one-dyad-out cross-validation and cross-testing methodologies. To detail: one session was allocated as the test set, while the remaining 16 sessions served as the training set. Within the training set, hyperparameters were optimized through 16 cross-validation folds. The final performance metrics were derived from the held-out dyad \#17. This process was systematically repeated for each of the 17 dyads, guaranteeing that every dyad was used as the test set exactly once. Although the dataset was relatively small, this procedure ensured a robust evaluation of the modeling technique while also allowing us to mitigate the risk of overfitting.

As in the evaluation of the large language models (LLMs) in \autoref{subsect:llmfusion}, each model was trained using all three input modalities, as well as through an ablation study involving various subsets of these modalities, detailed in \autoref{tab:rmse_prediction}. The results suggest that the Support Vector Machine (SVM) achieved the best performance across the majority of subsets, with the Random Forest (RF) model closely following. While these two models outperformed the LLM variants, they were the only models to do so: the remaining three models generally underperformed compared to the LLM variants.
\subsection{LLM Fusion\label{subsect:llmfusion}}

\texttt{GPT-4} was provided with the multimodal transcript paired with each of the survey items of the engagement questionnaire. Note that a few items on the questionnaire explicitly reference laughing or eye contact: despite not providing the model with explicit information on these behaviors, we included these items to explore the capability of the model to infer these behaviors with limited information. 

We performed a set of ablation experiments to explore the significance of various feature sets, notated with the following naming convention: \begin{itemize}
    \item \textbf{4}: This model was provided with the raw dialogue transcription alone.
    \item \textbf{S}: The transcription is preceded by participant survey responses to the personality and beliefs questionnaires as (\textbf{S})ystem instructions.
    \item \textbf{G}: The transcription is enhanced with descriptions of each participant's (\textbf{G})aze behavior during each speaking turn.
    \item \textbf{F}: The transcription is enhanced with descriptions of each participant's (\textbf{F})acial expression during each speaking turn.
\end{itemize}
In three instances, the length of the multimodal transcript with added descriptions exceeded the input constraints of \texttt{GPT-4} (two for \textbf{4SGF} and one for \textbf{4GF}); in these cases, the transcript was truncated. A $t$-test comparing the residuals of the truncated sessions with those of the non-truncated sessions yielded $p$-values of $0.186$ and $0.648$, suggesting no significant difference between the two groups. Future studies may benefit from exploring the impact of different truncation lengths and the ability of the technique to perform with shorter observation time.

The \emph{temperature} parameter was set at 0 to ensure sampling from the most likely responses to the questionnaire. In cases where \texttt{GPT-4} did not provide a numeric response, we selected the highest-likelihood numeric response from the top 20 generations for the first output token (see \autoref{appendix:llm_refusal} for more details on this process).

\subsubsection{LLM Fusion Results}

We evaluated this technique through two labeling tasks: predicting participants' exact responses and predicting the valence/arousal of their responses. The ``exact'' response refers to the specific answer given by participants (a numeric rating between 1 and 7). The valence/arousal model categorizes responses based on emotional dimensions: valence is defined as the positive or negative degree of emotion (e.g., pleasure/displeasure), and arousal is defined as the intensity of emotion (high or low)~(\cite{mollahosseini2019}). We define valence in terms of the ``disagree'' range, a score of 1 (``strongly disagree')' through 3 (``slightly disagree''), a neutral score of 4 (``neither agree nor disagree''), or the ``agree'' range, a score of 5 (``slightly agree'') through 7 (``strongly agree''). Arousal is calculated as the distance of the participants' rating from the neutral score of 4, i.e., $|\text{response}-4|$: e.g., ``strongly disagree'' and ``strongly agree'' would share the same arousal category.

\textbf{Exact Response}
As seen in \autoref{tab:llm_ablation}, \texttt{GPT-4}'s\footnote{We also experimented with \texttt{GPT-3.5}; however, given its significantly poorer performance against \texttt{GPT-4}, we elected to exclude its results from further analysis.} zero-shot performance of this technique is comparable to the baseline and classical early fusion models, evaluated via RMSE. Krippendorff's alpha metric, used to assess the reliability of agreement between multiple raters, indicated a moderate level of agreement between the model's predictions and the participants' responses, ranging within $[0.470, 0.543]$ across questions~(\cite{landis1977, wong2021}). This suggests that while the zero-shot technique may not outperform the more advanced models, it still holds potential for applications where computational resources are limited. Furthermore, the findings highlight the importance of evaluating various methodologies in diverse contexts, as different tasks may yield varying levels of effectiveness.

\textbf{Valence}
When restricting the labeling task to valence only, \texttt{GPT-4} predictions agree statistically significantly with the study's participants. As presented in \autoref{tab:llm_ablation}, all Krippendorff's alpha scores fall within an error interval of $[0.61, 0.80]$~(\cite{landis1977}). 

Upon closer inspection of the valence predictions of the \texttt{LLM-4S} ablation model (which achieved the strongest performance in labeling exact responses; presented in \autoref{appendix:llm_ablations}), we can observe that \texttt{GPT-4} reliably labels participant ``agree'' responses, with a class accuracy of $91.8\%$. However, \texttt{GPT-4} is less reliable in predicting participant ``disagree'' responses, achieving a class accuracy of $66.1\%$. Notably, \texttt{GPT-4} performs significantly poorly in labeling participant ``neutral'' responses, with a class accuracy of $12.7\%$. Given that the label range is smaller --- one possible value (4), as opposed to three values in ``agree'' (5, 6, 7) or ``disagree'' (1, 2, 3) ranges --- poor performance may be expected. We conjecture that \texttt{GPT-4}'s process of reinforcement learning from human feedback (RLHF)~(\cite{ouyang2022}) to reduce toxicity may result in overly ``positive'' responses from \texttt{GPT-4}, inadvertently introducing bias against ``negative'' responses. Further investigation is warranted to understand this potential for bias.

\textbf{Arousal}
Across all ablations, \texttt{GPT-4} performs poorly in labeling the arousal of a participant's response, only marginally better than chance given Krippendorff's alpha scores in the range of $[0.047, 0.071]$ (see \autoref{tab:llm_ablation} for detail). While \texttt{GPT-4} appears able to predict the general attitude of the participant towards a questionnaire statement (valence), it cannot reliably determine the strength of the participant's feelings (arousal).

\subsubsection{Contribution per modality}
To study the impact of each behavior modality (described in \ref{text_modalities}), we conducted a two-tailed paired $t$-test of each model's residuals against those of the baseline. The results suggest that each modality group added to the \textbf{LLM-4} baseline provides a statistically significant positive contribution ($p<0.05$) to model performance. 

\begin{table*}
\centering
\begin{tabular}{l||c|c|c}
\toprule
Ablation & Exact & Valence & Arousal \\ \midrule
4        & 0.470 (0.209)     & 0.634 (0.246)  &  0.055 (0.169)  \\
4S        & 0.518 (0.217)    &  0.687 (0.252) & \textbf{0.071 (0.174)}   \\
4F        & 0.513 (0.203)     & 0.686 (0.250) & 0.053 (0.164)   \\
4GF        & 0.520 (0.212)     & 0.695 (0.259) & 0.066 (0.180)  \\
4S        & \textbf{0.543 (0.206) }    & 0.680 (0.244)  & 0.054 (0.185)  \\ 
4SG        & 0.535 (0.210)     & 0.702 (0.247) & 0.039 (0.172)  \\
4SF        & 0.532 (0.202)     & 0.698 (0.247) & 0.055 (0.170)   \\
4SGF        & 0.531 (0.193)     & \textbf{0.703 (0.248)} & 0.047 (0.180)  \\
\bottomrule
\end{tabular}
\caption{\label{tab:llm_ablation}Krippendorff's alpha scores, mean and standard deviation, for each ablation (higher is better).}
\end{table*}

In contrast, the additional modalities worsen the performance of the \textbf{4S} baseline on labeling exact scores but improve the performance on labeling valence. In a paired $t$-test of residuals comparing exact predictions, the addition of facial expression descriptions in the \textbf{4SF} and \textbf{4SGF} ablations worsened performance significantly ($p=0.003$ and $p=0.021$, respectively); however, gaze did not have a notable impact ($p=0.164$).

\subsubsection{Performance across individual survey questions}
The following statements achieved the \emph{best} performance across all ablations (mean accuracy and standard deviation):
\begin{enumerate}[itemsep=0pt]
    \item \textit{I felt like my conversation partner really listened to me} (mean 64.0\%, std. dev. 7.1\%); 
    \item \textit{I became irritated with my partner at some points in the conversation} (mean 60.7\%, std. dev. 5.9\%); and 
    \item \textit{My conversation partner seemed like a warm person} (mean 53.7\%, std. dev. 6.2\%). 
\end{enumerate}

The following statements achieved the \emph{worst} performance across all ablations (mean accuracy and standard deviation):
\begin{enumerate}[itemsep=0pt]
    \item \textit{My conversation partner was quite sensitive} (mean 4.0\%, std. dev. 1.6\%);
    \item \textit{I would trust my conversation partner with sensitive information} (mean 8.8\%, std. dev. 5.2\%); and 
    \item \textit{My partner and I laughed during our interaction} (mean 10.3\%, std. dev. 4.1\%). 
\end{enumerate}

Prediction performance on the questions about laughter and eye contact is relatively poor, addressing our earlier hypothesis regarding the ability of the model to infer this behavior.  In general, while the transcript did not explicitly contain descriptions of laughter, \texttt{GPT-4} tends to respond with the assumption that laughter did occur. Although numerous caveats apply to these results, they generally reflect the opinions of our study's participants.

\section{Conclusion}

Engagement is fundamental to all human interactions, representing the intrinsic interest or emotional investment of the individuals involved. Despite humans' intuitive understanding of engagement, developing computational systems capable of recognizing and measuring engagement remains a significant challenge. Our work studies this core element of communication through smart glasses worn by participants in natural conversation. We collected a dataset of casual conversations between pairs of strangers, each outfitted with a pair of smart glasses, to capture behavioral cues such as facial expressions, eye contact, and verbal exchanges. We introduce a novel fusion method using large language models (LLMs), generating a ``multimodal transcript'' of the conversation to prompt an LLM to predict the participants' self-reported engagement levels. 
Our work is one of the first to use language to ``reason'' about human behavior, laying the groundwork for many promising directions for future research in computational behavior analysis.

However, it is crucial to acknowledge the limitations and biases associated with the models used. LLMs inadvertently learn and incorporate positional, racial, gender, and other social biases~(\cite{cheng2023, wan2023, navigli2023, wang2023b}). They are also sensitive to the wording of provided prompts.  Furthermore, given that our multimodal transcript relies on pre-trained models such as OpenFace, MediaPipe, and Whisper, possible issues of bias and robustness in those models \cite{namba2021viewpoint, graham2024evaluating} should also be taken into consideration. Additional noise may be created from the usage of multiple pre-trained models. The ability of the multimodal transcript to accurately represent the conversation is inherently limited by the accuracy of the pre-trained models used.

Given the limited size and variance in demographics of our participants and engagement experiences within our dataset, it also raises the question of how well LLMs can simulate engagement questionnaire responses for different populations and conversational experiences. It's also possible that a person's responses to the Big Five Inventory and belief questionnaire may not accurately reflect their true personality and beliefs.

LLMs such as \texttt{GPT-4} have been fine-tuned with RLHF to produce responses that are safer and better aligned with the user's intent. While this process reduces response toxicity and improves the ability to follow instructions, we note that this calibration may interfere with the ability of the LLM to emulate human-like responses in a research setting.
\vfill
\pagebreak
\bibliographystyle{ACM-Reference-Format}
\bibliography{refs}

\appendix
\clearpage
\begin{sidewaysfigure}
    \checkoddpage
    \ifoddpage
        \vspace{0.75\textwidth}
    \fi
    \raggedright
    \section{Engagement Questionnaire\label{appendix:questionnaire}}
    The following questionnaire was completed by each participant at the end of the recording session. Also displayed is the distribution of responses received in our participant sample; red rows indicate negatively-coded items.

    \vspace{0.5em}
    \hrule
    \vspace{0.5em}
    
    \textit{Please use this 7-point rating scale to share your impressions of the conversation with your partner.}

    \vspace{0.5em}
    
    \centering
    \resizebox{\textwidth}{!}{
    \input{fig/engagement_questionnaire/engagement_questionnaire.pgf}
    }
\end{sidewaysfigure}

\clearpage
\section{LLM Fusion: Valence Prediction\label{appendix:llm_ablations}}
One of the experiments described in \autoref{subsect:llmfusion} involves an evaluation of the LLM's ability to predict response valence rather than exact answers. Results for model \textbf{4S}, using raw transcripts and the participant personal characteristics, are presented in \autoref{tab:llm_confusion_valence}.
\begin{table*}
\centering
\begin{tabular}{l||ccc|c|c}
& \multicolumn{5}{c}{\textbf{Predicted}}\\
\midrule
\textbf{Actual} & \textbf{Agr.} & \textbf{Neu.} & \textbf{Dis.} & \textbf{All} & \textbf{Cl. Acc.} \\ \midrule
\textbf{Agr.}& 1072 & 44 & 52  & 1168 & 91.8 \\
\textbf{Neu.}& 91   & 18 & 33  & 142  & 12.7 \\
\textbf{Dis.}& 105  & 62 & 325 & 492  & 66.1 \\
\midrule
\textbf{All} & 1268 & 124 & 410 & 1802 & 56.9 \\
\bottomrule
\end{tabular}
\caption{\label{tab:llm_confusion_valence}4S Valence Prediction Confusion Matrix: Responses are categorized as (Dis)agree (1--3), (Neu)tral (4), or (Agr)ee (5--7). Class accuracy is also reported.}
\end{table*}

\section{LLM Fusion: Non-Numeric Responses \label{appendix:llm_refusal}}

There were 80 cases across all ablation experiments where the first generated token returned by the LLM was non-numeric ($1.1\%$ of all responses). For 57 of these questions, \texttt{GPT-4} declined to respond because it was prompted to answer questions that referenced modalities not explicitly provided in the ablation feature sets. Sample responses are included below for illustration.

\noindent Q8: \emph{My partner tended to make eye contact during the conversation.}
    \begin{itemize}
        \item ``\texttt{As an AI, I don't have the ability to make or observe eye contact.}''
        \item ``\texttt{As this conversation was text-based and not in person, I cannot accurately rate my partner's tendency to make eye contact. Therefore, I would rate it as N/A.}''
        \item ``\texttt{As this conversation was text-based and didn't involve any form of video or face-to-face in\-ter\-ac\-tion, I cannot rate my partner's eye contact. Therefore, I would rate it as N/A.}''
    \end{itemize}
\noindent Q47: \emph{My conversation partner made a lot of eye contact.}
    \begin{itemize}
        \item ``\texttt{As an AI, I don't have the ability to make or perceive eye contact. Therefore, I can't provide a rating for this statement.}''
        \item ``\texttt{As an AI, I don't have the ability to make or perceive eye contact. Therefore, I would rate this statement as 1, strongly disagree.}''
        \item ``\texttt{Given the nature of the conversation, it's hard to determine the level of eye contact as it was a text-based interaction. However, if we consider the level of engagement and attentiveness as a form of "eye contact" in this context, I}'' [\emph{text cut off}]
    \end{itemize}

\begin{table*}
    \centering
    
    \begin{tabular}{cc|cc@{\hspace{2em}}cc|c}
    
    & \multicolumn{1}{c}{\phantom{\small Unfortunately}} &&&&
    \multicolumn{1}{c}{\phantom{\small Unfortunately}} \\
    \cmidrule[\heavyrulewidth]{1-3}
    \cmidrule[\heavyrulewidth]{5-7}
    & Token & Prob. &&& Token & Prob. \\
    
    \cmidrule{1-3} \cmidrule{5-7}
    1 & \texttt{\small As} & 0.316 && 11 & \texttt{\small Sorry} & 0.002 \\
    2 & \texttt{\small [} & 0.283 && 12 & \texttt{\small Because} & 0.002 \\
    3 & \texttt{\small Since} & 0.214 && 13 & \texttt{\small The} & 0.001 \\
    4 & \texttt{\small I} & 0.104 && 14 & \texttt{\small 5} & 0.001 \\
    5 & \texttt{\small Given} & 0.042 && 15 & \texttt{\small 4} & 0.001 \\
    6 & \texttt{\small Considering} & 0.007 && 16 & \texttt{\small It} & 0.001 \\
    7 & \texttt{\small This} & 0.007 && 17 & \texttt{\small Without} & 0.001 \\
    8 & \texttt{\small Unfortunately} & 0.004 && 18 & \texttt{\small N} & 0.001 \\
    9 & \texttt{\small Ap} & 0.003 && 19 & \texttt{\small 3} & 0.001 \\
    10 & \texttt{\small Due} & 0.003 && 20 & \texttt{\small My} & 0.001 \\
    \cmidrule[\heavyrulewidth]{1-3}
    \cmidrule[\heavyrulewidth]{5-7}
    
    \end{tabular}
    \caption{\label{tab:sample_dist}Sample top 20 tokens from a questionnaire response by the LLM where the first response is non-numeric.}
\end{table*}
For example, consider the following response to Q8: ``\texttt{As this conversation was text-based, I cannot provide a rating for eye contact}''. A sample of the top 20 tokens with highest probability are displayed in \autoref{tab:sample_dist}.

The other 23 responses exceeded 50 generated tokens and were cut off. This occurred often in the \textbf{4F} ablation experiments when the \texttt{GPT-4} would prefix its answers with the facial expression string, such as the following example.

\noindent ``\texttt{[You] [You are speaking mostly with relaxed facial muscles, a straight mouth, a smooth forehead, and un\-re\-mark\-able eyebrows.
Your partner is listening to you mostly with relaxed facial muscles, a straight mouth, a smooth forehead, and unremark}'' [\emph{text cut off}].

It's interesting to note that not all \texttt{GPT} models are able to impersonate a participant. For example, nearly all experiments with \texttt{gpt-4-1106-preview} would result in an example similar to the following:

\noindent``\texttt{As an AI language model, I don't have personal ex\-pe\-ri\-ences or opinions. However, if I were to simulate a response for the scenario described where a participant has engaged in an interesting conversation that touched on computer science, philosophy of neuroscience, dif\-fer\-ences between cities, and personal experiences, they might rate the conversation on the higher end of the scale indicating that they found it to be engaging and intellectually stimulating.}''
\newpage
\begin{figure*}[h]
\raggedright
    \section{Multimodal Transcript Template\label{appendix:llm_template}}
    This appendix contains a detailed version of the sample multimodal transcript depicted in \autoref{fig:teaser_figure}. {\color{Magenta}Magenta text} corresponds to information from personal inventories. {\color{red}Red text} corresponds to information from OpenFace. {\color{Plum}Violet text} corresponds to information from MediaPipe and Pupil Invisible eye tracking. {\color{blue}Blue text} corresponds to information from the Whisper transcription. {\color{PineGreen}Green text} corresponds to information from the post-session engagement survey. Black text is always present. The last row with ``assistant" is what the LLM generates.

    \vspace{1em}

    \centering
    \begin{tabular}{l|p{0.7\paperwidth}}
    \toprule
    Role & Content \\
    \midrule
    System & \texttt{You are a student at ... You are participating in a psychology study that aims to understand how people communicate, and you are participating in a conversation with ... as part of this study. There will be a questionnaire at the end of this conversation. Others will read what you answer; your goal is to convince them it was answered from the perspective of the persona that participated in the following conversation. }
    
    \texttt{\color{Magenta}Your personality traits are defined by the scores to the following statements. The scores range from 1 to 5, where 1 means strongly disagree and 5 means strongly agree. }
    
    \texttt{\textcolor{Magenta}{\qquad[Alice's personality defined by responses to the big-5 personality survey.]}}
    
    \texttt{\textcolor{Magenta}{Your political beliefs are defined by the following statements:}}
    
    \texttt{\textcolor{Magenta}{\qquad[Alice's beliefs defined by responses to the beliefs survey.]}}\\
    \midrule
    Assistant & \texttt{[You]}

    \texttt{[\textcolor{Plum}{You are looking at your partner's face about 80\% of the time.}}

    \texttt{\textcolor{red}{\qquad You are speaking with a smiling mouth, raised cheeks...}}
    
    \texttt{\textcolor{Plum}{\qquad Your partner is looking at your face about 80\% of the time.}}
    
    \texttt{\textcolor{red}{\qquad Your partner is listening with relaxed facial expression...}]}
    
    \texttt{\textcolor{blue}{Hi, I'm Alice! What year are you?}}\\
    \midrule
    User & \texttt{[Partner]}
    
    \texttt{[\textcolor{Plum}{You are looking at your partner's face about 60\% of the time.}}

    \texttt{\textcolor{red}{\qquad You are listening with a smiling mouth, raised cheeks...}}
    
    \texttt{\textcolor{Plum}{\qquad Your partner is looking at your face about 80\% of the time.}}
    
    \texttt{\textcolor{red}{\qquad Your partner is speaking with a smiling mouth, raised cheeks...}]}
    
    \texttt{\textcolor{blue}{Hi Alice, I'm Bob. I'm a sophomore.}} \\
    
    \midrule
    \rule{0pt}{4ex}
    &\rule[-3ex]{0pt}{0ex}[\textit{five minutes of conversation}]\\
    \midrule
    User & \texttt{[Experimenter] \textcolor{PineGreen}{On a scale of 1 to 7, where 1 means strongly disagree and 7 means strongly agree, how would you rate the following statement given the conversation you just had?}}
    
    \texttt{\textcolor{PineGreen}{I found this conversation to be interesting.}}
    
    \texttt{Your answers will be kept private and your conversation partner will not see the responses, so please be as honest as possible. Provide your answer in the form of an integer between 1 and 7.}\\
    \midrule
    Assistant & \texttt{7} \\
    \bottomrule
    \end{tabular}
\end{figure*}
\newpage

\clearpage
\section{Belief Questionnaire\label{appendix:belief}}
Each participant completed the following questionnaire at the end of the recording session. 

\vspace{0.5em}
\hrule
\vspace{1em}

\emph{Please select the answer which most represents your beliefs.}
\begin{multicols*}{2}
\hrule
\vspace{0.5em}

{\textbf{Environmental Protection}
\begin{itemize}[leftmargin=!, labelindent=1em, itemindent=-1em, itemsep=0pt]
\item I am very much against environmental protection.
\item I am against environmental protection.
\item I am mildly against environmental protection.
\item I am mildly in favor of environmental protection.
\item I am in favor of environmental protection.
\item I am very much in favor of environmental protection.

\end{itemize}
\vspace{0.5em}}

\hrule
\vspace{0.5em}

{\textbf{Careers for Women}
\begin{itemize}[leftmargin=!, labelindent=1em, itemindent=-1em, itemsep=0pt]
\item I am very much against women pursuing careers.
\item I am against women pursuing careers.
\item I am mildly against women pursuing careers.
\item I am mildly in favor of women pursuing careers.
\item I am in favor of women pursuing careers.
\item I am very much in favor of women pursuing careers.

\end{itemize}
\vspace{0.5em}}

\hrule
\vspace{0.5em}

{\textbf{Belief in God}
\begin{itemize}[leftmargin=!, labelindent=1em, itemindent=-1em, itemsep=0pt]
\item I strongly believe that there is a God.
\item I believe there is a God.
\item I feel that perhaps there is a God.
\item I feel that perhaps there is no God.
\item I believe there is no God.
\item I strongly believe there is no God.

\end{itemize}
\vspace{0.5em}}

\hrule
\vspace{0.5em}

{\textbf{Ranking of Schools}
\begin{itemize}[leftmargin=!, labelindent=1em, itemindent=-1em, itemsep=0pt]
\item I am very much against the ranking of schools.
\item I am against the ranking of schools.
\item I am mildly against the ranking of schools.
\item I am mildly in favor of the ranking of schools.
\item I am in favor of the ranking of schools.
\item I am very much in favor of the ranking of schools.

\end{itemize}
\vspace{0.5em}}
\hrule
\vfill\mbox{}
\columnbreak
\hrule
\vspace{0.5em}

{\textbf{Abortion}
\begin{itemize}[leftmargin=!, labelindent=1em, itemindent=-1em, itemsep=0pt]
\item I am very much against abortion.
\item I am against abortion.
\item I am mildly against abortion.
\item I am mildly in favor of abortion.
\item I am in favor of abortion.
\item I am very much in favor of abortion.

\end{itemize}
\vspace{0.5em}}
\hrule
\vspace{0.5em}

{\textbf{Death Penalty}
\begin{itemize}[leftmargin=!, labelindent=1em, itemindent=-1em, itemsep=0pt]
\item I am very much against the death penalty.
\item I am against the death penalty.
\item I am mildly against the death penalty.
\item I am mildly in favor of the death penalty.
\item I am in favor of the death penalty.
\item I am very much in favor of the death penalty.

\end{itemize}
\vspace{0.5em}}

\hrule
\vspace{0.5em}

{\textbf{Gay Marriage}
\begin{itemize}[leftmargin=!, labelindent=1em, itemindent=-1em, itemsep=0pt]
\item I am very much against gay marriage.
\item I am against gay marriage.
\item I am mildly against gay marriage.
\item I am mildly in favor of gay marriage.
\item I am in favor of gay marriage.
\item I am very much in favor of gay marriage.

\end{itemize}
\vspace{0.5em}}
\hrule
\vspace{0.5em}

{\textbf{Money}
\begin{itemize}[leftmargin=!, labelindent=1em, itemindent=-1em, itemsep=0pt]
\item I strongly believe that money is one of the most important things in life.
\item I believe that money is one of the most important things in life.
\item I feel perhaps that money is one of the most important things in life.
\item I feel perhaps that money is not one of the most important things in life.
\item I believe that money is not one of the most important things in life.
\item I strongly believe that money is not one of the most important things in life.

\end{itemize}
\vspace{0.5em}}
\hrule
\vfill\mbox{}
\columnbreak
\hrule
\vspace{0.5em}

{\textbf{Divorce}
\begin{itemize}[leftmargin=!, labelindent=1em, itemindent=-1em, itemsep=0pt]
\item I am very much against divorce.
\item I am against divorce.
\item I am mildly against divorce.
\item I am mildly in favor of divorce.
\item I am in favor of divorce.
\item I am very much in favor of divorce.

\end{itemize}
\vspace{0.5em}}
\hrule
\vspace{0.5em}

{\textbf{Smoking}
\begin{itemize}[leftmargin=!, labelindent=1em, itemindent=-1em, itemsep=0pt]
\item I am very much against smoking in public places like bars.
\item I am against smoking in public places like bars.
\item I am mildly against smoking in public places like bars.
\item I am mildly in favor of smoking in public places like bars.
\item I am in favor of smoking in public places like bars.
\item I am very much in favor of smoking in public places like bars.

\end{itemize}
\vspace{0.5em}}
\hrule
\vspace{0.5em}

{\textbf{Spanking Children}
\begin{itemize}[leftmargin=!, labelindent=1em, itemindent=-1em, itemsep=0pt]
\item In general, I am very much in favor of spanking children.
\item In general, I am in favor of spanking children.
\item In general, I am mildly in favor of spanking children.
\item In general, I am mildly against spanking children.
\item In general, I am against spanking children.
\item In general, I am very much against spanking children.

\end{itemize}
\vspace{0.5em}}
\hrule
\vspace{0.5em}

{\textbf{Climate Change}
\begin{itemize}[leftmargin=!, labelindent=1em, itemindent=-1em, itemsep=0pt]
\item I strongly believe that climate change has not been accelerated by humans.
\item I believe that climate change has not been accelerated by humans.
\item I mildly believe that climate change has not been accelerated by humans.
\item I mildly believe that climate change has been accelerated by humans.
\item I believe climate change has been accelerated by humans.
\item I strongly believe that climate change has been accelerated by humans.

\end{itemize}
\vspace{0.5em}}
\hrule
\vfill\mbox{}
\columnbreak
\hrule
\vspace{0.5em}

{\textbf{Health Care}
\begin{itemize}[leftmargin=!, labelindent=1em, itemindent=-1em, itemsep=0pt]
\item I strongly believe that humans are not entitled to health care.
\item I believe that humans are not entitled to health care.
\item I mildly believe that humans are not entitled to health care.
\item I mildly believe that humans are entitled to health care.
\item I believe that humans are entitled to health care.
\item I strongly believe that humans are entitled to health care.

\end{itemize}
\vspace{0.5em}}

\hrule
\vspace{0.5em}

{\textbf{Social Safety Net}
\begin{itemize}[leftmargin=!, labelindent=1em, itemindent=-1em, itemsep=0pt]
\item I strongly believe the government should not provide funds to support individuals' welfare.
\item I believe the government should not provide funds to support individuals' welfare.
\item I mildly believe the government should not provide funds to support individuals' welfare.
\item I mildly believe the government should provide funds to support individuals' welfare.
\item I believe the government should provide funds to support individuals' welfare.
\item I strongly believe the government should provide funds to support individuals' welfare.

\end{itemize}
\vspace{0.5em}}

\hrule
\vspace{0.5em}

{\textbf{College}
\begin{itemize}[leftmargin=!, labelindent=1em, itemindent=-1em, itemsep=0pt]
\item I strongly believe the government should not pay for college students' tuition.
\item I believe the government should not pay for college students' tuition.
\item I mildly believe the government should not pay for college students' tuition.
\item I mildly believe the government should pay for college students' tuition.
\item I believe the government should pay for college students' tuition.
\item I strongly believe the government should pay for college students' tuition.

\end{itemize}
\vspace{0.5em}}
\hrule
\vfill\mbox{}
\columnbreak
\hrule
\vspace{0.5em}

{\textbf{[Local University]}
\begin{itemize}[leftmargin=!, labelindent=1em, itemindent=-1em, itemsep=0pt]
\item I strongly believe that [local university] is a welcoming university environment.
\item I believe that [local university] is a welcoming university environment.
\item I mildly believe that [local university] is a welcoming university environment.
\item I mildly believe that [local university] is not a welcoming university environment.
\item I believe that [local university] is not a welcoming university environment.
\item I strongly believe that [local university] is not a welcoming university environment.

\end{itemize}
\vspace{0.5em}}

\hrule
\end{multicols*}

\end{document}